\documentclass{article}
\date{}
\usepackage[english]{babel}

\usepackage[letterpaper,top=2cm,bottom=2.5cm,left=3cm,right=3cm,marginparwidth=1.75cm]{geometry}

\usepackage{setspace}
\usepackage{amssymb,amsmath,amsthm}
\usepackage{graphicx}
\usepackage[table,xcdraw]{xcolor}
\usepackage{multirow} 
\usepackage{caption}
\usepackage{subcaption}
\usepackage[colorlinks=true, allcolors=blue]{hyperref}
\usepackage{rotating}
\usepackage{adjustbox}
\usepackage[normalem]{ulem}
\useunder{\uline}{\ul}{}
\usepackage{lscape}
\usepackage{longtable}

\title{A Comprehensive Framework for Reliable Legal {AI}: Combining Specialized Expert Systems and Adaptive Refinement}

\usepackage{authblk}
\author[1]{Sidra Nasir\thanks{corresponding author: sidranasir.rajput@univr.it}}
\author[2]{Qamar Abbas}
\author[3]{Samita Bai}
\author[4]{Rizwan Ahmed Khan}

\affil[1]{Dipartimento di Informatica, Università di Verona, Italy}

\affil[2]{Faculty of Information Technology, Salim Habib University, Karachi, Pakistan}

\affil[3]{Canadian Institute of Cybersecurity (CIC), University of New Brunswick, Fredericton, New Brunswick, Canada}

\affil[4]{Department of Computer Science, School of Mathematics and Computer Science, Institute of Business Administration, Karachi, Pakistan}

\begin{document}
\maketitle

\begin{abstract}

This article discusses the evolving role of artificial intelligence (AI) in the legal profession, focusing on its potential to streamline tasks such as document review, research, and contract drafting. However, challenges persist, particularly the occurrence of "hallucinations" in AI models, where they generate inaccurate or misleading information, undermining their reliability in legal contexts. To address this, the article proposes a novel framework combining a mixture of expert systems with a knowledge-based architecture to improve the precision and contextual relevance of AI-driven legal services. This framework utilizes specialized modules, each focusing on specific legal areas, and incorporates structured operational guidelines to enhance decision-making. Additionally, it leverages advanced AI techniques like Retrieval-Augmented Generation (RAG), Knowledge Graphs (KG), and Reinforcement Learning from Human Feedback (RLHF) to improve the system's accuracy. The proposed approach demonstrates significant improvements over existing AI models, showcasing enhanced performance in legal tasks and offering a scalable solution to provide more accessible and affordable legal services. The article also outlines the methodology, system architecture, and promising directions for future research in AI applications for the legal sector.

\end{abstract}

\section{Introduction}
The legal profession is undergoing a significant transformation, driven in part by advances in AI \cite{sun2023short}. As AI becomes increasingly adept at understanding and generating human language, its potential for streamlining complex legal processes is becoming evident. Tasks such as reviewing legal documents, conducting research, and drafting contracts can now be supported or automated with the help of AI tools~\cite{padiu2024extent}. However, the integration of AI into legal applications presents challenges, including the tendency of AI models, especially large language models (LLMs), to produce misleading or incorrect information, a phenomenon known as "hallucinations" \cite{magesh2024hallucination, wiratunga2024cbr}. Ethical AI in legal applications ensures fairness, transparency, and accountability, aligning AI-driven decisions with established legal standards and human-centric values to enhance trust and reliability~\cite{nasir2024ethical}.

These errors arise because AI models generate responses based on patterns identified in their training data rather than on verified legal facts \cite{rawte2023survey}. While this allows for the creation of fluent and coherent text, it also introduces the risk of inaccuracies that can have serious legal consequences. In an industry where precision and reliability are paramount, such mistakes can significantly undermine the effectiveness of AI solutions in legal contexts.

Simultaneously, there is a growing global demand for affordable legal services. Traditional legal support is often prohibitively expensive and time-consuming~\cite{lai2024large}, leaving many individuals without adequate representation. AI holds the promise of making legal assistance more accessible and scalable, particularly for routine tasks like document analysis and preliminary consultations. This potential underscores the critical need for AI systems that are not only efficient but also highly accurate and reliable.

To address the issues of inaccuracy and enhance AI’s applicability in legal services, we propose a novel framework that incorporates a \textbf{mixture of expert systems} alongside a \textbf{knowledge-based architecture}. This framework strategically allocates legal tasks to specialized modules, or "experts," each designed to handle specific areas of law. By utilizing focused, domain-specific knowledge, the system ensures that legal advice and analysis are more precise and contextually relevant compared to general-purpose AI models. Additionally, a \textbf{knowledge integration system} underpins the framework by providing access to reliable legal information, thereby reducing the likelihood of errors.

Beyond the architectural enhancements, our framework implements \textbf{structured operational guidelines} modeled on real-world legal practices. These guidelines emulate the methodologies employed by law firms, ensuring that the AI follows a systematic approach in tasks such as information gathering, analysis of legal precedents, and formulation of conclusions. This structured methodology helps prevent the compounding of errors, thereby improving the overall quality and reliability of the AI-generated legal outputs.

Our approach demonstrates significant improvements over existing models, particularly in handling the complexities of legal language and logical reasoning. Compared to other AI systems, our model achieves better results in legal benchmarking tests and real-world case studies, highlighting its ability to provide more dependable legal advice. By combining specialized expert knowledge with a structured operational framework, this solution paves the way for more accurate, efficient, and accessible legal services.

In the following sections, the paper discusses key aspects of its proposed framework. The Introduction sets the stage by highlighting challenges in integrating AI into the legal profession, such as inaccuracies and the importance of reliable systems. The Literature Review, refer Section \ref{LR}, explores current AI techniques in the legal domain, focusing on large language models, domain-specific methods, and approaches to minimize hallucinations. The Methodology section, refer Section \ref{Method}, presents the novel framework, detailing components like Retrieval-Augmented Generation, Knowledge Graphs, a Mixture of Experts approach, and Reinforcement Learning from Human Feedback. In Section \ref{Res} we have discussed Results and  highlighted the framework's performance improvements over existing models in legal tasks. Finally Section \ref{Conc}, the Conclusion, summarizes findings and suggests directions for future research in advancing reliable legal AI systems.





\section{\textsc{Literature Review}} \label{LR}

Large Language Models (LLMs), such as GPT-3 \cite{brown2020language} and GPT-4 \cite{achiam2023gpt}, have marked significant advancements in natural language processing (NLP) through their transformer-based architecture. These models have demonstrated exceptional performance across various text generation tasks, excelling in areas like text summarization and question answering (Q/A) \cite{shaghaghian2020customizing}. However, the success of these LLMs largely hinges on training data from general-purpose domains. This becomes problematic when applying them to specialized fields like law, which is characterized by intricate language, domain-specific terminology, and nuanced logic.

To bridge this gap, researchers have turned to fine-tuning LLMs specifically for legal tasks. By adapting pre-trained models with legal data, LLMs can better capture the complexities of legal texts \cite{chalkidis2020legal}. The process of fine-tuning allows these models to utilize their inherent strengths in language generation while integrating the specialized knowledge required for tasks such as contract analysis, statutory interpretation, and case law reasoning. This literature review critically examines the existing techniques employed for major NLP tasks in the legal domain, with a focus on text summarization and question-answering. 

Key areas of exploration include:

\begin{itemize}
	\item The potential benefits and limitations of fine-tuning LLMs for legal applications.
	\item Promising and cost-effective methods for successfully adapting LLMs.
	\item Emerging research gaps and future directions for advancing legal NLP.
\end{itemize}

The development of Large Language Models (LLMs) has significantly transformed natural language understanding across diverse sectors, including finance \cite{yang2023fingpt} and medicine \cite{helwan2023medical}. Despite these advancements, applying LLMs within the legal domain presents unique challenges. General-purpose LLMs exhibit robust generalization capabilities; however, their effectiveness diminishes when deployed in specialized fields such as law \cite{yue2023disc}.

To address these challenges, recent research has focused on fine-tuning LLMs specifically for legal applications. For example, ChatLaw \cite{cui2023chatlaw}, an open-source LLM customized for the legal sector, was fine-tuned using the Low-Rank Adaptation (LoRA) technique \cite{hu2021lora} on a comprehensive legal dataset. This fine-tuning process aimed to reduce hallucinations—instances where models generate plausible yet factually incorrect information—by integrating vector database retrieval with keyword-based retrieval. Such an approach enhances accuracy, particularly during legal consultations.

In the context of Chinese legal systems, similar progress has been observed. Yue et al. \cite{yue2023disc} fine-tuned the Baichuan-13B-Base model using the DISC-Law-SFT dataset, which encompasses various legal tasks like judgment prediction and document summarization. By incorporating authentic legal texts, including judicial verdicts and regulations, this methodology ensures that the model produces contextually relevant and grounded outputs, thereby validating the effectiveness of domain-specific fine-tuning for legal applications.

Additionally, LexGPT \cite{lee2023lexgpt} represents another significant effort, wherein the model was fine-tuned on the Pile of Law dataset \cite{henderson2022pile} for legal classification tasks. However, empirical results indicated that even with fine-tuning, existing models lag behind discriminative models specifically designed for legal tasks. This discrepancy highlights the limitations of current fine-tuning techniques, particularly for applications demanding high precision and sophisticated legal reasoning.

The preference for fine-tuning over training LLMs from scratch is both practical and economically advantageous, as emphasized by Yang et al. \cite{yang2023fingpt}. Training a legal-specific LLM requires substantial computational resources, whereas fine-tuning an existing model on a high-quality legal dataset offers a more efficient and scalable alternative.

While fine-tuning remains the predominant strategy for optimizing LLMs for legal tasks, prompting techniques to provide a complementary approach. Prompt-based methods involve guiding the model with specific instructions or examples, which can significantly enhance performance on tasks requiring complex reasoning or logical argumentation. For instance, Yu et al. \cite{yu2022legal} investigated Chain-of-Thought (CoT) prompting, a technique that directs models through multi-step reasoning processes. Although CoT prompting showed promise in improving legal reasoning tasks, the study revealed that LLMs still struggle to emulate the deep logical structures inherent in legal decision-making. Similarly, Blair et al. \cite{blair2023can} evaluated various prompting techniques on the SARA dataset \cite{holzenberger2020dataset}, finding that while GPT-3 performed well on basic legal tasks, its effectiveness diminished in more complex scenarios.

These findings highlight the critical need for domain-specific models tailored for legal reasoning. Despite their inherent flexibility, general-purpose LLMs may require additional fine-tuning or specialized prompts to achieve dependable outcomes in the legal arena.

Moreover, the integration of Reinforcement Learning with Human Feedback (RLHF) has emerged as a promising method for addressing issues such as hallucinations in LLMs \cite{rawte2023survey}. By incorporating human evaluations into the training regimen, RLHF enhances the model’s performance, ensuring that outputs are more aligned with factual accuracy and user requirements. For example, TacticalGPT \cite{carontacticalgpt} utilized RLHF to improve decision-making in sports analytics by incorporating expert feedback. In the legal domain, RLHF can be instrumental in producing reliable and factually accurate outputs. Legal-specific LLMs like ChatLaw \cite{cui2023chatlaw} have begun to integrate RLHF, thereby further reducing hallucinations and making these models more suitable for high-stakes legal applications.

Despite these advancements, several gaps persist in the development of legal LLMs. Although fine-tuning has demonstrated potential, the persistence of hallucinations indicates that more rigorous methodologies are necessary to ensure reliability. A significant barrier is the scarcity of high-quality, domain-specific datasets, which hampers the performance of current models. Additionally, achieving explainability within the complex legal domain remains a formidable challenge.

Furthermore, as discussed in studies such as \cite{cui2023chatlaw}, the trustworthiness of legal LLMs remains questionable, especially when outputs are not easily verifiable. Researchers must continue exploring methods that integrate retrieval-based systems and explainability tools to enhance the transparency and reliability of legal LLMs. Ensuring that legal outputs are both accurate and comprehensible to non-experts is a critical challenge that must be addressed to advance the application of LLMs in the legal field.

\section{Methodology} \label{Method}

In artificial intelligence, particularly within fields such as \textbf{Generative AI} and \textbf{Multi-Modal Applications}, the ability to generate accurate and contextually relevant responses to complex queries is essential. Our system utilizes the combined strengths of \textbf{Retrieval-Augmented Generation (RAG)} \cite{lewis2020retrieval} and \textbf{Knowledge Graphs (KG)} \cite{li2024simple}, further enhanced by a \textbf{Mixture of Experts (MoE)} \cite{ge2024openagi} framework and \textbf{Reinforcement Learning from Human Feedback (RLHF)} \cite{christiano2017deep}, to deliver precise and contextually enriched answers. This integration ensures that the system not only retrieves pertinent information but also comprehends and utilizes the intricate relationships within the data to generate informed and continuously improving responses.

\subsection{Datasets Selection}
A range of tasks was selected, aligned with the specified capability levels to support a balanced assessment across various dimensions. The tasks and their associated details are summarized in Table~\ref{tab:Tab1}. In cases where multiple datasets were available for a given task, the most recent version was chosen to maintain relevance. This approach aims to provide a structured basis for evaluation across the selected tasks.

\begin{table}[!ht]
	\resizebox{\columnwidth}{!}{%
		\begin{tabular}{lllll}
			\hline
			\textbf{Expert}          & \textbf{Task}            & \textbf{Dataset} & \textbf{Size}                    & \textbf{Metric}   \\ \hline
			Consultant               & Question Answering        & LegalQA~\cite{askari2024answer}          & 9,000+ questions, 33,000 answers & Accuracy          \\ \hline
			\multirow{4}{*}{Researcher} & Cases Identification & CaseHold~\cite{zheng2021does} \& LEDGAR~\cite{tuggener2020ledgar} &53,000+ \& 60,000 contracts & Rouge-L \\
			& Article Recitation       & LEXTREME~\cite{katz2024gpt}            & 11 datasets                  & Rouge-L           \\
			& Element Extraction       &  \& COLIEE~\cite{goebel2023summary}       &  13,000+ documents                & F1 Score \\
			& Text Classification               & SARA~\cite{holzenberger2020dataset} \& LexGlue~\cite{chalkidis2020legal}            & 867 questions, 768 statutes      & Accuracy          \\ \hline
			\multirow{2}{*}{Paralegal}  & Document Summarization   & Billsum~\cite{kornilova2019billsum}          & 18,000 summaries                 & BLEU              \\
			& Contract Drafting        & CUAD~\cite{hendrycks2021cuad}             & 13,000+ cases                    & Rouge-L           \\ \hline
			\multirow{2}{*}{Advisor} & Case Analysis            & Super-SCOTUS\cite{fang2023super}~\& EUR-LEX~\cite{aumiller2022eur}          & 5,205~\& 55,000 cases                    & Rouge-L, Accuracy \\
			
			& Judgment Prediction      & ECHR~\cite{chalkidis2019neural}             & 11,500 cases                     & Rouge-L           \\ \hline
		\end{tabular}%
	}
	\caption{Details of the dataset}
	\label{tab:Tab1}
\end{table}

\subsection{System Architecture}

The architecture of our system is meticulously designed to facilitate seamless interaction between RAG, Knowledge Graphs, the MoE framework, and the RLHF process. When a user submits a query \( x \) through the \textbf{User and Expert Interface}, the system initiates a multi-stage processing workflow. The query \( x \) is first transformed into a dense vector representation \( \mathbf{v}_x \) using an embedding function \( f \), enabling efficient similarity computations. Concurrently, the system accesses a \textbf{Knowledge Graph (KG)}, a structured repository of interconnected data points, to enrich the retrieval process with relational context.

\subsection{Retrieval-Augmented Generation (RAG) Module with Knowledge Graph Integration}

At the core of our system lies the \textbf{Retrieval-Augmented Generation (RAG)} module, which orchestrates the retrieval and generation of responses to complex queries. The RAG model comprises two primary components: the \textbf{Retriever} and the \textbf{Generator}. The Retriever maps both the input query \( x \) and potential documents \( d \in \mathcal{K} \) from the knowledge base to their respective vector representations \( \mathbf{v}_x \) and \( \mathbf{v}_d \) using the embedding function \( f \) (LegalBERT). The similarity between \( x \) and each document \( d \) is quantified using cosine similarity, calculated as:

\begin{equation}
	\text{sim}(x, d) = \frac{\mathbf{v}_x \cdot \mathbf{v}_d}{\|\mathbf{v}_x\| \|\mathbf{v}_d\|}
\end{equation}

To ensure high precision in retrieval, we apply a similarity threshold \( \theta \), which is empirically tuned based on legal document validation. Typically, \( \theta \) is set between 0.8 and 0.9, allowing only highly relevant documents to pass, and was chosen based on empirical validation with legal documents. This range was found to strike a balance between recall and precision, allowing the retrieval of only highly relevant documents while filtering out less relevant ones. Legal texts tend to be verbose, and high \( \theta \)values help maintain precision, crucial for high-stakes applications.:

\begin{equation}
	D = \{d_i \in \mathcal{K} \mid \text{sim}(x, d_i) \geq \theta\}
\end{equation}

These retrieved documents \( D \) provide the foundational information that the \textbf{Generator} utilizes to formulate a coherent and contextually relevant response \( y \). The Generator operates on a Transformer-based architecture, modeling the conditional probability distribution:

\begin{equation}
	P(y | x, D) = \prod_{t=1}^{T} P(y_t | y_{<t}, x, D)
\end{equation}

where each token \( y_t \) in the response is generated based on the preceding tokens \( y_{<t} \), the original query \( x \), and the retrieved documents \( D \).

\subsubsection{\textit{Knowledge Graph Facilitation in RAG}}

The \textbf{Knowledge Graph (KG)} plays a pivotal role in enhancing the RAG process by providing a structured and relational context that enriches both the retrieval and generation phases. 

We utilize techniques such as Named Entity Recognition (NER) to identify legal entities like statutes, case names, organizations, and legal terms. Relation Extraction determines the types of relationships between these entities, such as "cites," "applies to," "overruled by," and "related to." Coreference Resolution ensures consistency in entity representation by resolving different references to the same entity. Dependency Parsing and Semantic Role Labeling (SRL) analyze the grammatical and semantic structures of sentences to understand how entities interact within the text. Part-of-Speech (POS) Tagging provides foundational linguistic information that supports these analyses. Extracted entities and relationships are structured into triplets.

The KG is formally defined as:
\begin{equation}
	KG = \{(h, r, t) \mid h \text{ is the head entity}, r \text{ is the relation}, t \text{ is the tail entity}\}
\end{equation}

In this triplet structure: where \( h \) (head entity) is the starting point of the relationship. \( r \) (relation) defines the connection or association between the two entities. \( t \) (tail entity) is the endpoint of the relationship. When a query \( x \) is processed, the system not only retrieves relevant documents based on textual similarity but also explores the relational data within the KG to identify interconnected entities and concepts that provide deeper context.

Consider a legal Knowledge Graph that includes statutes, legal cases, and principles. For instance, a triplet might be represented as:

\begin{equation}
	(\text{Statute X}, \text{applies\_to}, \text{Contract Law})
\end{equation}

\begin{itemize}
	\item \( h \): \emph{Statute X} represents a specific law or regulation.
	\item \( r \): \emph{applies\_to} indicates the nature of the connection between the two entities, showing that the statute is relevant to a particular area of law.
	\item \( t \): \emph{Contract Law} represents the legal domain or area impacted by the statute.
\end{itemize}

This relational information is embedded into the retrieval process, allowing the Retriever to access not just documents that match the query terms but also those that are contextually relevant through their connections in the KG. Mathematically, the similarity measure is augmented to incorporate relational context:

\begin{equation}
	\text{sim}(x, d) = \frac{\mathbf{v}_x \cdot \mathbf{v}_d + \alpha \cdot \text{sim}(KG_x, KG_d)}{\|\mathbf{v}_x\| \|\mathbf{v}_d\| + \alpha}
\end{equation}

Where \( \text{sim}(KG_x, KG_d) \) represents entity and relation similarity in the KG, using techniques like TransE embeddings, and \( \alpha \) balances KG contributions, typically set to 0.5.  These embeddings are computed offline, meaning each entity and relationship has a fixed vector representation.

The integration of the Knowledge Graph with the RAG module significantly enhances the retrieval process. Instead of relying solely on unstructured documents, the Retriever can now access structured, relational data that provides deeper insights into the query context. This means that the system can identify not only the most relevant documents based on textual similarity but also understand the relationships and connections between different entities within those documents.

\begin{equation}
	\text{sim}(x, d) = \beta \cdot \text{sim}_{\text{text}}(x, d) + (1 - \beta) \cdot \text{sim}_{\text{KG}}(x, d)
\end{equation}

Where, $\text{sim}_{\text{text}}(x,d)$ represents the cosine similarity between the query and document embeddings. $\text{sim}_{\text{KG}}(x,d)$ denotes the similarity based on Knowledge Graph embeddings or graph-based measures. The parameter $\beta$ balances the importance of textual similarity and KG similarity.

During the generation phase, the enriched information from the Knowledge Graph ensures that the Generator can produce responses that are not only contextually appropriate but also factually accurate and legally compliant. By employing the relational data from the KG, the Generator gains a more comprehensive understanding of the query's context, enabling it to generate more informed and precise responses.

For instance, consider a legal query: "What precedent cases support the application of statute X in contract disputes?" The system retrieves documents related to statute X and contract disputes, enriched with relational data from the KG such as related cases, involved parties, and relevant legal principles. This comprehensive retrieval allows the Generator to construct a response that not only lists the precedent cases but also explains their relevance and contextual connections to statute X and contract disputes.

\begin{figure}[!ht] \centering \includegraphics[width=15cm, height=21cm]{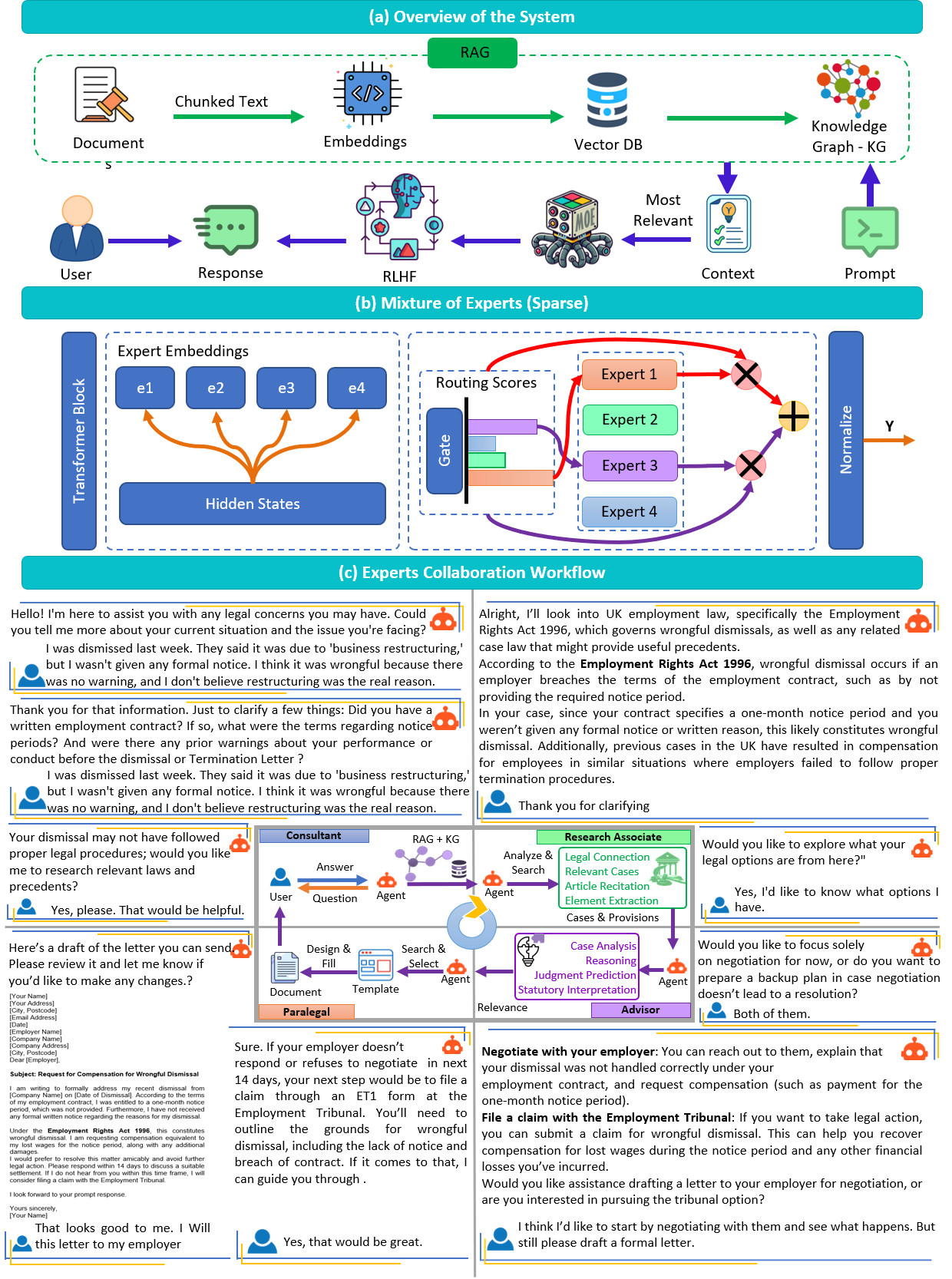} \caption{A comprehensive AI-assisted legal system integrating Retrieval-Augmented Generation (RAG) (a), sparse Mixture of Experts (MoE)(b), and a collaborative workflow involving consultants, research associates, Paralegal, and advisors for accurate legal analysis and decision-making (c).} \label{fig:3} \end{figure}

\subsection{\textbf{Mixture of Experts (MoE) Framework and Experts Collaboration Workflow}}

To handle the diversity and complexity inherent in legal queries, our system incorporates a \textbf{sparse Mixture of Experts (MoE)} framework. This framework dynamically allocates computational resources by routing inputs to specialized experts tailored to specific sub-domains within the legal field. The MoE framework consists of a collection of specialized models \( \{E_1, E_2, \dots, E_N\} \), each fine-tuned on distinct sub-domains.

Each agent \( E_i \) within the MoE model is specialized to handle specific legal tasks, such as case classification, article recitation, reasoning, legal element extraction, and case analysis. When a consultant presents \( k \) questions to the system, each related to different aspects \( Q = \{q_1, q_2, \dots, q_k\} \), the \textbf{Gating Network} evaluates each query \( q_i \) and assigns routing probabilities \( \mathbf{g} = [g_1, g_2, \dots, g_N] \) to each expert based on their relevance to the query:

\begin{equation}
	\mathbf{g} = \text{GatingNetwork}(x; \theta_{\text{gate}})
\end{equation}

Ensuring that the sum of the gating probabilities equals one:

\begin{equation}
	\sum_{i=1}^{N} g_i = 1
\end{equation}

The gating network determines which experts are most relevant to the input query 
$x$ by outputting a probability distribution over all experts.

\begin{equation}
	\mathbf{g} = \text{softmax}\left( \mathbf{W}_{\text{gate}} \cdot \mathbf{v}_x + \mathbf{b}_{\text{gate}} \right)
\end{equation}

To maintain computational efficiency, only the top \( K \) experts with the highest routing scores are activated:

\begin{equation}
	\mathcal{E}(x) = \text{TopK}(\mathbf{g}, K) = \{E_i \mid i \in \text{indices of top } K \text{ scores in } \mathbf{g}\}
\end{equation}

Each activated expert \( E_i \) processes the input independently, producing an output \( \mathbf{h}_i \):

\begin{equation}
	\mathbf{h}_i = E_i(q_i; \theta_i) \quad \forall E_i \in \mathcal{E}(x)
\end{equation}

The outputs from these experts are then normalized and aggregated to form the final MoE output \( \mathbf{h}_{\text{MoE}} \):

\begin{equation}
	\mathbf{h}_{\text{MoE}} = \sum_{i=1}^{N} g_i \cdot E_i(q_i; \theta_i) \quad \text{subject to} \quad g_i = 0 \text{ for } i \notin \mathcal{E}(x)
\end{equation}

This aggregated output \( \mathbf{h}_{\text{MoE}} \) is seamlessly integrated into the RAG generation pipeline, enhancing the response with specialized knowledge from relevant sub-domains. The integration ensures that the system not only retrieves and generates responses based on general information but also utilizes domain-specific expertise to address nuanced legal queries effectively.

A critical aspect of our system is the \textbf{Experts Collaboration Workflow}, which bridges human expertise with AI-driven insights through \textbf{Reinforcement Learning from Human Feedback (RLHF)}. This workflow involves structured interactions among various roles, ensuring that AI-generated responses meet high standards of accuracy and relevance while continuously improving based on human feedback.

The collaborative process begins with the \textbf{Consultant}, who initiates case analysis by defining objectives and outlining specific requirements. The Consultant translates high-level requirements into structured queries \( x \), ensuring that the system is tasked with well-defined objectives. These queries \( x \) are then processed by the \textbf{Research Associate}, who conducts in-depth legal research, extracts pertinent case information, and interacts with the RAG model to retrieve and interpret relevant documents \( D \). This role ensures that the retrieved information aligns with legal standards and contextual relevance.

Each question \( q_i \) from the Consultant is routed to the appropriate expert \( E_j \) within the MoE architecture, where:

\begin{equation}
	E_j(q_i) = \text{Process}\leftarrow \text{Domain Specific}(q_i)
\end{equation}

The \textbf{Advisor} utilizes the MoE-enhanced RAG model for statutory interpretations and case analyses. Combining AI-driven predictions with human judgment, the Advisor performs tasks such as assignment prediction and element extraction. The outputs from the experts \( \{E_j(q_i)\} \) are aggregated:

\begin{equation}
	y_{\text{aggregated}} = \sum_{i=1}^{k} E_j(q_i)
\end{equation}

This aggregated output \( y_{\text{aggregated}} \) is forwarded to the Advisor for review. Upon approval, the \textbf{Paralegal} finalizes the document, ensuring its compliance with legal standards and preparing release templates. The Paralegal reviews and refines AI-generated outputs \( y \), providing feedback \( \delta \) to the system. The entire collaborative process can be represented as:

\begin{equation}
	y_{\text{final}} = \text{Paralegal} \left( \text{Advisor} \left( \sum_{i=1}^{k} E_j(q_i) \right) \right)
\end{equation}

\subsection {Reinforcement Learning from Human Feedback (RLHF)}

\textbf{Reinforcement Learning from Human Feedback (RLHF)} is an integral component of our system, ensuring that the AI continuously learns and adapts based on human expertise and preferences. RLHF enhances the \textbf{Experts Collaboration Workflow} by providing a structured method for incorporating human feedback into the model's training process, thereby improving response quality over time and aligning the AI’s responses more closely with human judgments and legal standards. By incorporating direct feedback from experts, RLHF ensures that the system's outputs are not only accurate but also contextually appropriate and compliant with evolving legal practices.

The process begins with \textbf{Feedback Collection}, where after the Paralegal reviews the AI-generated response \( y \), feedback \( \delta \) is gathered. This feedback can be either numeric (e.g., a rating or score) or qualitative (e.g., textual comments). Qualitative feedback is first converted to numeric values using a sentiment or rating scale; for instance, "high relevance" might correspond to a score of 1.0, while "low relevance" could be scored as 0.5. To account for feedback type and quality, the weight \( w_k \) associated with each feedback component \( \delta_k \) is adjusted dynamically, ensuring that more reliable or domain-specific feedback has a higher impact on the model’s learning.

In the subsequent step of \textbf{Reward Modeling}, the collected feedback \( \delta \) is used to construct a reward signal \( R \) that quantifies the quality of the response. Positive feedback enhances the reward, while negative feedback reduces it:

\begin{equation}
	R = \sum_{k} w_k \cdot \delta_k
\end{equation}

Where \( w_k \) are dynamically assigned weights that reflect the importance or reliability of different aspects of the feedback \( \delta_k \).

During the \textbf{Policy Optimization} phase, the system employs reinforcement learning algorithms, specifically \textbf{Proximal Policy Optimization (PPO)}, to adjust the model parameters \( \theta \) in order to maximize the expected reward \( R \):

\begin{equation}
	\theta \leftarrow \theta + \eta \cdot \nabla_\theta \mathbb{E}[R]
\end{equation}

Here, \( \eta \) is the learning rate, and \( \nabla_\theta \mathbb{E}[R] \) represents the gradient of the expected reward with respect to the model parameters. PPO is chosen for its stability and efficiency in handling sparse rewards, making it particularly suitable for legal applications where high-quality feedback (or high rewards) may be less frequent but highly informative. Additionally, PPO’s stable policy updates prevent abrupt shifts in model behavior, maintaining consistent response quality—a critical aspect in legal contexts where erratic changes could undermine reliability and user trust.
The \textbf{Iterative Improvement} process in RLHF involves updating the model after every substantial batch of feedback is collected, typically every 100-200 interactions, or when feedback metrics indicate a performance plateau. In our system, feedback metrics serve as quantitative indicators of the AI's output quality, guided by expert feedback across several key dimensions. These include the \emph{Relevance Score}, which measures how closely the AI’s response aligns with the context and specifics of the query, ensuring that the response contains pertinent information. The \emph{Accuracy Rating} assesses factual correctness, particularly vital in legal contexts to prevent errors in interpreting case law, statutes, or legal principles. Additionally, a \emph{Compliance Check} metric evaluates whether the AI-generated responses adhere to established legal standards and regulations, confirming their appropriateness within legal frameworks. Finally, a \emph{User Satisfaction Score} reflects the overall usefulness, clarity, and coherence of the AI’s response, providing insights into user experience quality. Monitoring these metrics enables the system to gauge performance effectively and implement updates based on a substantial sample size, enhancing response quality without frequent, disruptive changes. By aggregating feedback in meaningful batches, the system makes stable parameter adjustments, fostering consistent improvements aligned with user expectations and legal standards.

\section{Results and Discussion}\label{Res}
This section presents and analyzes the performance outcomes of our proposed system, which integrates Retrieval-Augmented Generation (RAG), Knowledge Graphs (KG), a Mixture of Experts (MoE) framework, and Reinforcement Learning from Human Feedback (RLHF). By evaluating large language models (LLMs) such as GPT-4 \cite{achiam2023gpt}, LLaMA-3 \cite{dubey2024llama}, and Google Flan-T5 \cite{chung2024scaling} across various legal tasks, we demonstrate how different architectures and methodologies impact performance in specialized fields like law.

\begin{figure}[hbt!] 
	\centering \includegraphics[width=9cm, height=13cm]{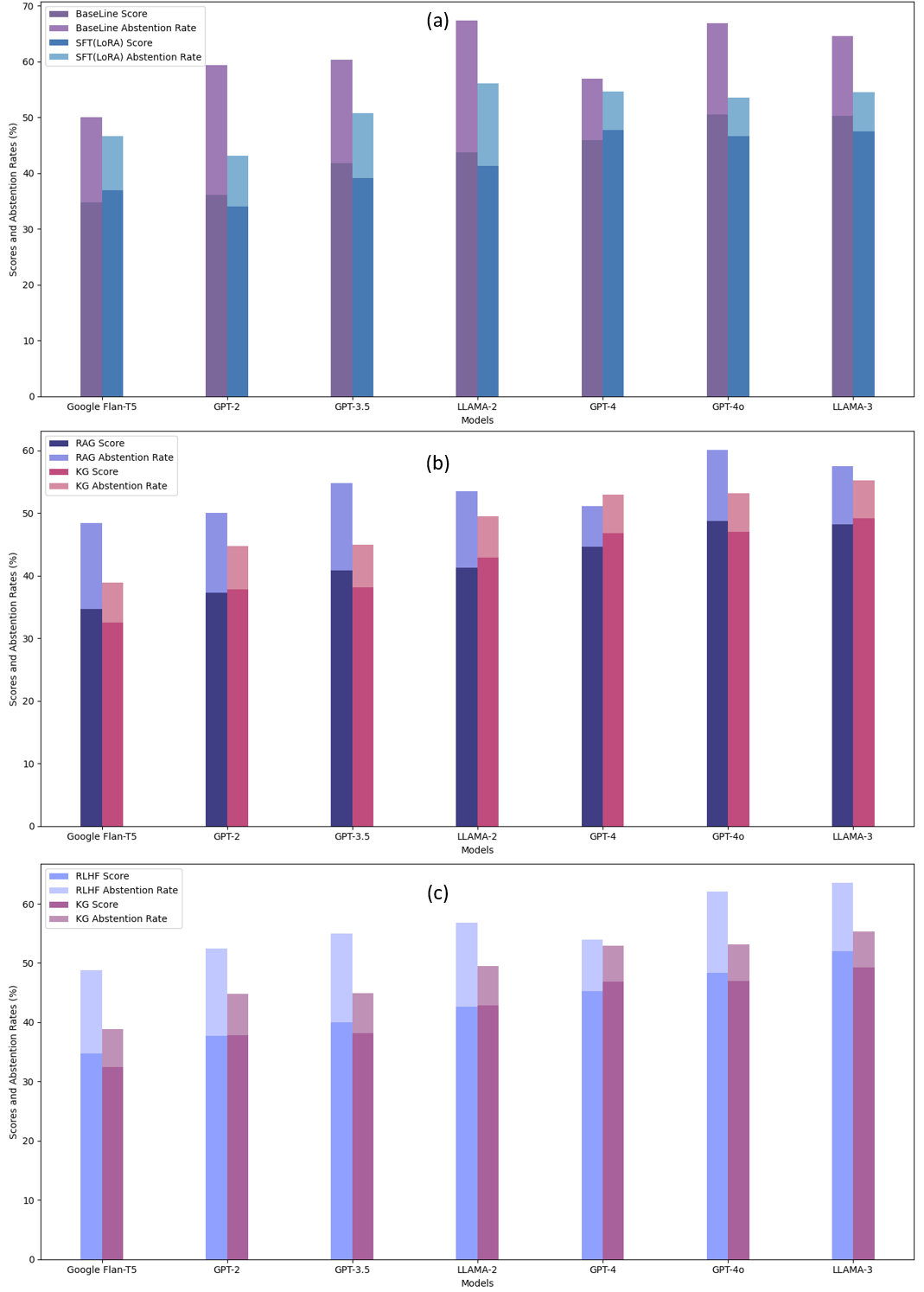} 
	\caption{Comparative Analysis of Model Enhancements ((a) Baseline vs SFT(LoRA), (b)RAG vs KG and KG vs RLHF) Across Architectures} 
	\label{fig:4} 
\end{figure}

\begin{figure}[hbt!] 
	\centering \includegraphics[width=12cm, height=8cm]{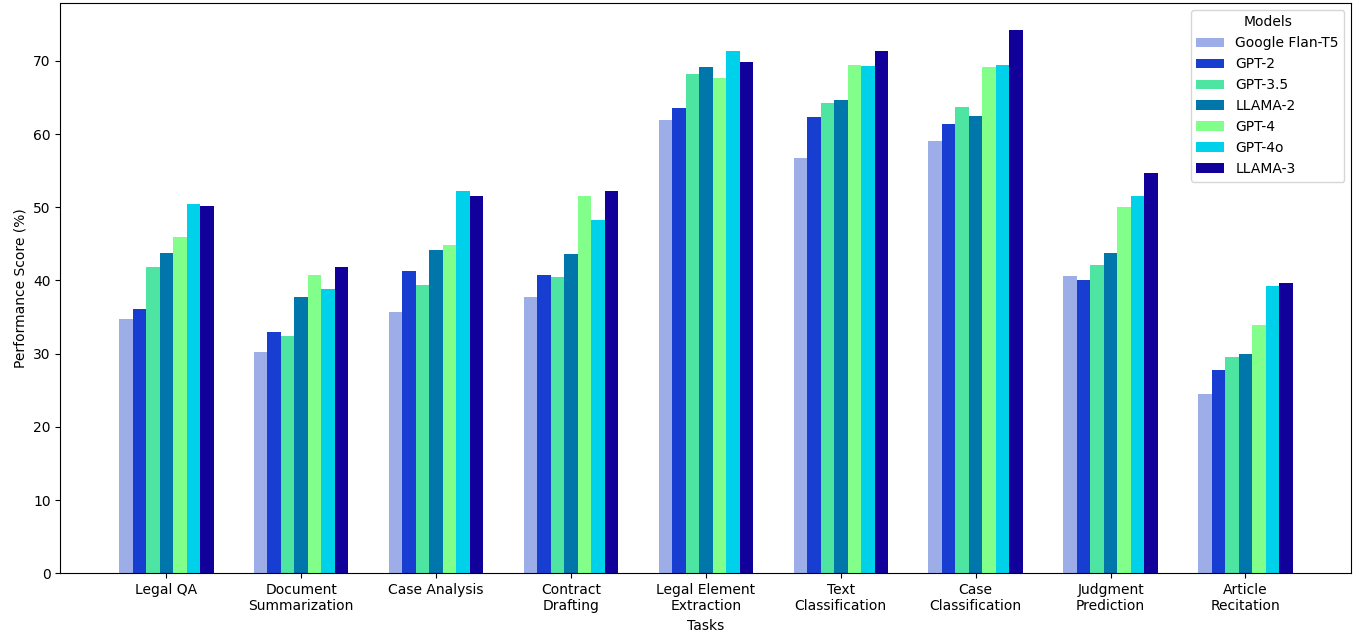} 
	\caption{Task-Wise Performance Comparison of Different Models Across Legal Applications.} 
	\label{fig:6} 
\end{figure}

Figure~\ref{fig:6} (a) showcases the comparative performance between baseline models and those fine-tuned using Supervised Fine-Tuning (SFT) with Low-Rank Adaptation (LoRA). The data indicates a substantial improvement in task scores across all models when employing SFT-LoRA. For instance, LLAMA-3's performance escalates from approximately 40\% in the baseline to 60\% post fine-tuning. This 20\% increase exhibits the efficacy of the SFT-LoRA approach in enhancing model capabilities.

Moreover, the reduction in abstention rates by around 15\% suggests that fine-tuning not only boosts accuracy but also equips models to respond more confidently. The advanced models like GPT-4 and LLAMA-3 exhibit better adaptability to fine-tuning, outperforming older architectures such as GPT-2 and Google Flan-T5, which show limited gains. This disparity highlights the importance of modern, adaptable architectures for fine-tuning processes to achieve optimal performance enhancements.

However, despite these improvements, fine-tuned models occasionally struggled with handling complex and contextually rich queries. This limitation necessitated the integration of additional strategies, such as RAG and RLHF, to further enhance performance and reliability.

Figure~\ref{fig:6} (b)compares the performance and abstention rates between models enhanced with Retrieval-Augmented Generation (RAG) and those integrated with Knowledge Graphs (KG). RAG-enhanced models generally achieve higher scores and lower abstention rates compared to KG-enhanced setups. For instance, GPT-4 and LLAMA-3 show substantial improvements with RAG, achieving scores around 60\%, while abstention rates are maintained below 20\%. The RAG framework benefits from its ability to retrieve contextually relevant information and generate detailed responses, making it particularly effective in complex, unstructured tasks such as judgment prediction and document summarization.

The performance of RAG is consistent across all models, although the extent of improvement varies. LLaMA-3, for example, showcases the best overall balance between task scores and abstention rates by employing RAG to integrate unstructured information effectively. In contrast, older architectures such as GPT-2 and Flan-T5 struggle to fully capitalize on RAG due to their limited contextual comprehension capabilities. This indicates that while RAG significantly enhances performance in advanced models, its benefits are constrained in less sophisticated architectures.

Figure~\ref{fig:6} (c) examines the performance and abstention rates between models enhanced with Reinforcement Learning from Human Feedback (RLHF) and those integrated with Knowledge Graphs (KG). RLHF consistently outperforms KG integration in most unstructured and complex tasks, such as judgment prediction and document summarization. For example, GPT-4 achieves an accuracy of approximately 65\% with RLHF, compared to 55\% with KG integration. This 10\% improvement signifies RLHF's effectiveness in aligning model outputs with human expectations, particularly in open-ended scenarios where nuanced understanding is crucial.
\begin{figure}[!htbp] 
	\centering \includegraphics[width=10cm, height=8cm]{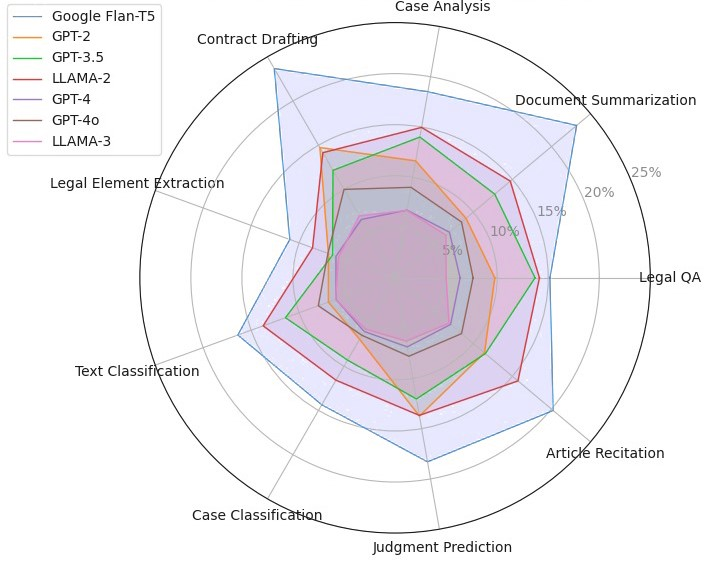} 
	\caption{Performance comparison of various language models across nine legal tasks based on abstention rate.} 
	\label{fig:5} 
\end{figure}

Conversely, KG integration excels in structured tasks like legal element extraction and text classification. The relational data embedded within KGs allows models to utilize interconnected information effectively, resulting in competitive scores despite slightly higher abstention rates (ranging from 25-30\%) in more complex tasks. This indicates that while RLHF is more effective for general and open-ended tasks, KG integration provides significant benefits for tasks requiring structured and relational understanding. Therefore, combining RLHF with KG integration emerges as a promising hybrid approach, harnessing the strengths of both methodologies to mitigate their limitations.

Figure~\ref{fig:5} offers a comparative radar chart of different models across multiple tasks, emphasizing their respective strengths and weaknesses. LLAMA-3 and GPT-4 consistently achieve the highest scores across most tasks, demonstrating their versatility and robustness. Their balanced performance in both structured tasks (e.g., text classification) and unstructured tasks (e.g., case analysis and document summarization) highlights their adaptability and the effectiveness of the fine-tuning and RLHF strategies applied.

In contrast, models such as GPT-2 and Flan-T5 exhibit limited capabilities, particularly in tasks requiring complex reasoning or semantic understanding. This performance gap highlights the limitations of older architectures in handling advanced, nuanced legal queries. The radar chart also reveals that no single model excels uniformly across all tasks, suggesting that selecting or customizing models based on specific task requirements is crucial. Employing hybrid approaches that combine the strengths of different models could further improve overall performance, ensuring that specialized tasks receive appropriate attention.

Figure~\ref{fig:5} illustrates that GPT-4 outperforms other models like GPT-2, GPT-3.5 \cite{brown2020language}, LLaMA-2 \cite{touvron2023llama}, LLaMA-3 \cite{dubey2024llama}, and Google Flan-T5 \cite{chung2024scaling} across nine legal tasks. This consistent top performance indicates GPT-4's strong capability to understand and generate complex legal texts, making it a reliable tool for legal professionals requiring accurate and contextually relevant outputs.

Tasks like article recitation pose challenges for all models, reflected in uniformly lower scores. This difficulty likely arises from the need for precise retrieval and verbatim reproduction, areas where generative models inherently struggle. On the other hand, structured tasks such as legal element extraction consistently yield high scores across all models, demonstrating their proficiency in handling clear, well-defined input-output relationships. This suggests that integrating retrieval-based and generative approaches can provide a more robust framework for managing a diverse array of tasks.

Figure~\ref{fig:6} compares performance scores across various tasks, including Legal QA, Document Summarization, Case Analysis, Contract Drafting, Legal Element Extraction, Text Classification, Case Classification, Judgment Prediction, and Article Recitation, for different models (Flan-T5, GPT-2, GPT-3.5, LLAMA-2, GPT-4, GPT-4o, LLAMA-3).

Tasks such as Legal Element Extraction, Text Classification, and Case Classification demonstrate consistently high performance across all models, with LLAMA-3 and GPT-4 achieving the highest scores (~70\%). These tasks are more structured, making them easier for the models to process and complete effectively. Tasks like Document Summarization, Judgment Prediction, and Article Recitation show lower performance across the board. For instance, Article Recitation has the lowest scores (~40-50\%), reflecting the inherent difficulties in handling verbatim reproduction and retrieval. Judgment Prediction also highlights variability among models, with RLHF-enhanced models such as GPT-4 and LLAMA-3 performing better than earlier architectures like GPT-2 and Flan-T5.

\begin{figure}[hbt!] 
	\centering \includegraphics[width=15cm, height=8cm]{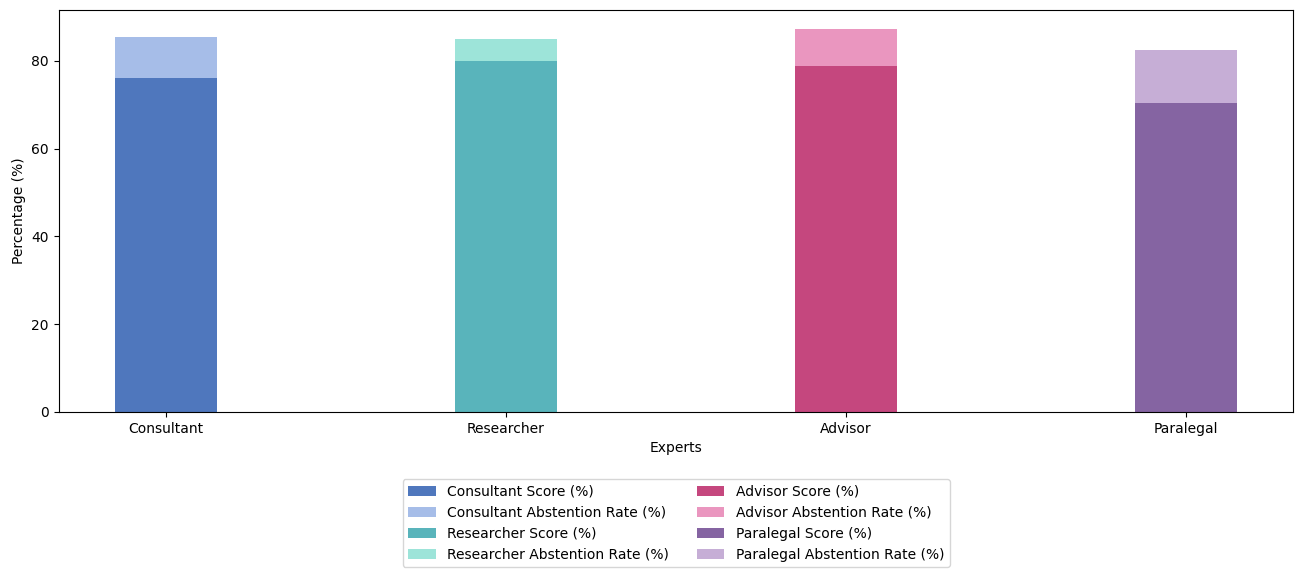} 
	\caption{Role-Based Task Performance Evaluation Across Expert Roles (Consultant, Researcher, Advisor, Paralegal).} 
	\label{fig:7} 
\end{figure}

Figure~\ref{fig:7} evaluates the performance and abstention rates of models in simulating the roles of Consultant, Researcher, Advisor, and Paralegal. The Consultant and Researcher roles achieve the highest scores, with task performance exceeding 80\% accuracy. This reflects the structured and well-defined nature of tasks assigned to these roles, such as information retrieval and case classification. These roles also exhibit low abstention rates (~15-20\%), indicating the models’ confidence in generating responses.

In contrast, the Advisor and Paralegal roles face greater challenges due to the complexity of tasks such as legal summarization and judgment prediction. These tasks require deeper reasoning and contextual understanding, resulting in slightly lower scores (~70\%) and higher abstention rates (~30\%). The Advisor role, in particular, struggles with tasks demanding long-range dependencies and nuanced interpretations, highlighting the need for further optimization in handling complex decision-making processes.

Supervised Fine-Tuning with Low-Rank Adaptation (SFT-LoRA) enhances model accuracy and confidence in structured tasks. For more complex, context-rich queries, integrating Retrieval-Augmented Generation (RAG) and Knowledge Graphs (KG) improves performance by reducing hallucinations and providing relational understanding. Advanced models like GPT-4 and LLaMA-3 benefit most from these integrations, outperforming older architectures in unstructured tasks such as judgment prediction and document summarization. Additionally, Reinforcement Learning from Human Feedback (RLHF) aligns outputs with human expectations, lowering abstention rates and ensuring appropriate responses. Combining RLHF with KG integration in hybrid approaches optimizes contextual reasoning and relational comprehension, thereby enhancing overall system performance and reliability.

The results affirm the pivotal role of integrated methodologies in advancing AI-driven legal applications. The synergistic combination of RAG, KG, MoE, and RLHF fosters a robust framework capable of handling diverse and complex legal queries with enhanced accuracy and confidence. The reasoning behind employing different strategies stems from the observed limitations of fine-tuning alone, particularly in managing hallucinations, contextual irrelevance, and toxicity. By integrating RAG, the system grounds its responses in reliable data sources, thereby reducing hallucinations and ensuring contextual relevance. RLHF, implemented via Proximal Policy Optimization (PPO), further refines the model by incorporating human feedback to avoid toxic outputs and align responses with human expectations.

\section{Conclusion} \label{Conc}

This research presents the development of an AI-driven legal assistant that enhances the precision and dependability of legal services through the integration of advanced artificial intelligence methodologies. Central to our approach is the implementation of a \textbf{Mixture-of-Experts (MoE)} framework combined with a \textbf{multi-agent collaborative} architecture. This design strategically allocates specialized legal tasks—such as contract analysis, statutory interpretation, and case prediction—to distinct expert models that have been fine-tuned for their respective domains. This setup ensures that each legal query is processed with the necessary specificity and accuracy, effectively mitigating the inaccuracies and hallucinations commonly associated with general-purpose large language models (LLMs).

A key component of our system is the incorporation of a \textbf{Knowledge Graph (KG)}-enhanced \textbf{Retrieval-Augmented Generation (RAG)} module. This integration uses structured data from extensive legal repositories to ground generated responses in verified legal information. The RAG module facilitates the retrieval of contextually relevant data, which is then used to produce detailed and factually accurate outputs. Additionally, the deployment of \textbf{Reinforcement Learning with Human Feedback (RLHF)} allows for continuous model refinement, ensuring alignment with established legal standards and practitioner requirements. The empirical evaluation across nine distinct legal tasks demonstrates that our system outperforms existing AI models. Performance metrics such as Rouge-L, BLEU, and F1 scores indicate significant enhancements in critical areas like legal question answering, document summarization, and case analysis. These results validate the effectiveness of our approach in combining specialized expert models with structured knowledge integration and iterative human-in-the-loop feedback mechanisms.

The modular and extensible nature of our system offers substantial flexibility for future advancements. Future work will explore the expansion of the expert modules to cover additional legal domains, further integration of real-time legal updates into the knowledge base, and the enhancement of explainability features to provide clearer insights into the AI's decision-making processes. Additionally, ongoing collaboration with legal professionals will be essential to refine the system's capabilities and ensure its practical applicability in real-world legal settings.

\addcontentsline{toc}{chapter}{References}
\bibliographystyle{ieeetr} 


\end{document}